\documentclass[letterpaper]{article} 
\usepackage{aaai2026}  
\usepackage{times}  
\usepackage{helvet}  
\usepackage{courier}  
\usepackage[hyphens]{url}  
\usepackage{graphicx} 
\usepackage{natbib}  
\usepackage{caption} 
\usepackage{algorithm}
\usepackage{algorithmic}

\usepackage{newfloat}
\usepackage{listings}
\DeclareCaptionStyle{ruled}{labelfont=normalfont,labelsep=colon,strut=off} 
\floatstyle{ruled}
\newfloat{listing}{tb}{lst}{}
\floatname{listing}{Listing}
\usepackage{amssymb}
\usepackage{amsmath}
\usepackage{mathrsfs}

\usepackage{graphicx} 
\usepackage{booktabs} 
\usepackage{array}    
\usepackage{caption}  
\usepackage{multirow}
\usepackage{amsmath}   
\usepackage{xcolor}    
\usepackage{array}     

\usepackage{pgfplots}

\usepackage{graphicx} 
\usepackage{subfigure} 

\graphicspath{{./images/}} %

\title{No Need for Real 3D: Fusing 2D Vision with Pseudo 3D Representations for Robotic Manipulation Learning}
\author{
    Run Yu,
    Yangdi Liu,
    Wen-Da Wei,
    Chen Li
}
\affiliations{
    HuaZhong University of Science and Technology\\

\detokenize{run-yu@hust.edu.cn}
    
}

\usepackage{bibentry}

\begin{document}

\maketitle

\begin{figure*}[t]
    \centering
    \includegraphics[width=\textwidth]{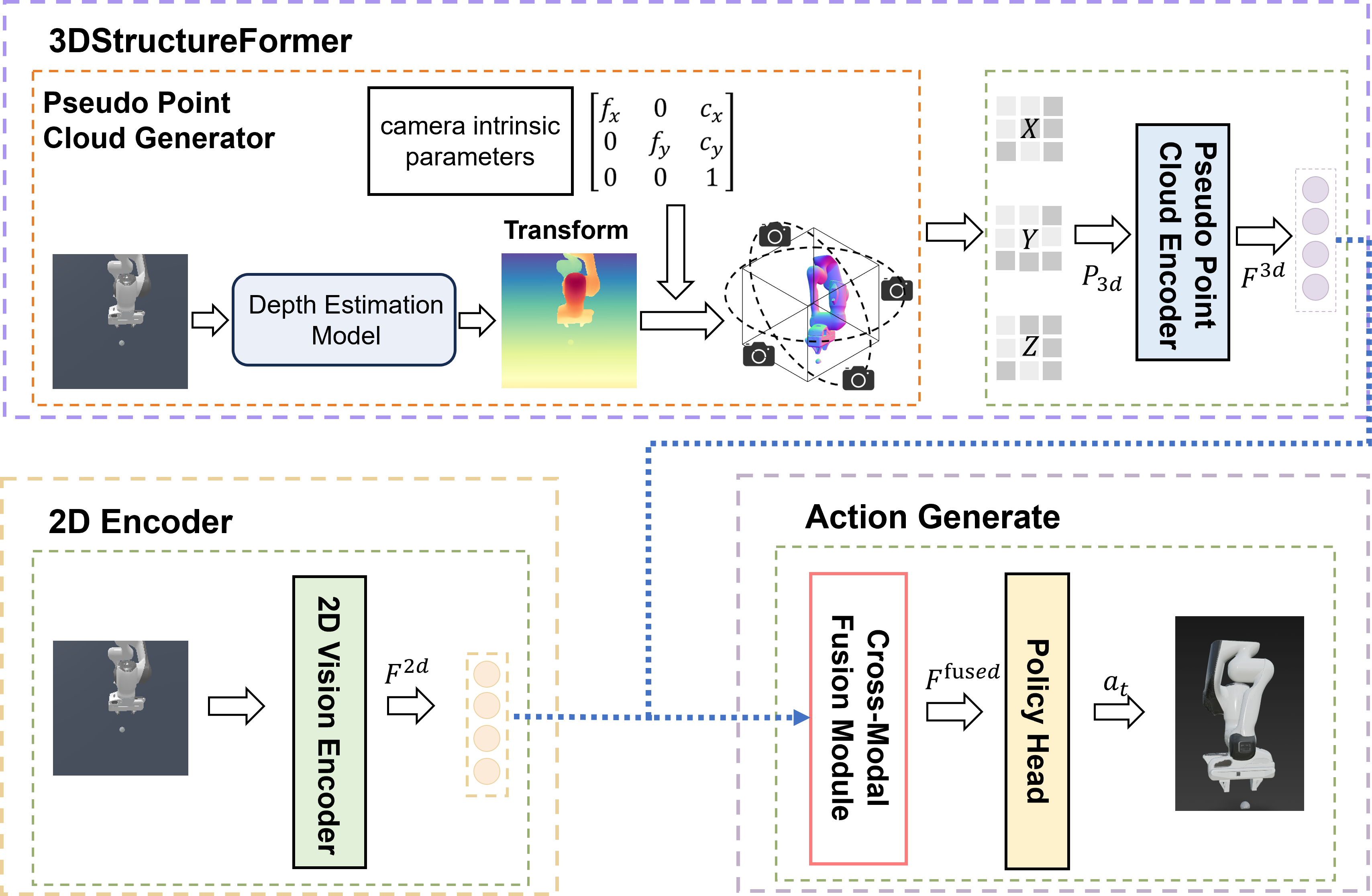} 
    \caption{Overview of NoReal3D. Our framework takes a monocular RGB image $\mathbf{I}$ as input. The 3DStructureFormer module, comprising a Pseudo Point Cloud Generator and a Pseudo Point Cloud Encoder, extracts geometrically meaningful pseudo-point cloud features $\mathbf{F}^{\text{3d}}$. These 3D features are fused with 2D visual features (extracted by a standard vision backbone like ViT or ResNet) to create a 3D-enhanced fused representation $\mathbf{F}^{\text{fused}}$. This fused representation improves spatial understanding and geometric perception for robotic manipulation tasks, enabling plug-and-play enhancement of existing 2D foundation models without requiring real 3D input or modifying the backbone architecture.}
    \label{fig:pipeline}
\end{figure*}

\begin{abstract}
Recently, vision-based robotic manipulation has garnered significant attention and witnessed substantial advancements.
2D image-based and 3D point cloud-based policy learning represent two predominant paradigms in the field, with recent studies showing that the latter consistently outperforms the former in terms of both policy performance and generalization,  thereby underscoring the value and significance of 3D information.
However, 3D point cloud-based approaches face the significant challenge of high data acquisition costs, limiting their scalability and real-world deployment. 
To address this issue, we propose a novel framework \textbf{NoReal3D}: which introduces the 3DStructureFormer, a learnable 3D perception module capable of transforming monocular images into geometrically meaningful pseudo-point cloud features, effectively fused with the 2D encoder output features.
Specially, the generated pseudo-point clouds retain geometric and topological structures so we design a pseudo-point cloud encoder to preserve these properties, making it well-suited for our framework.
We also investigate the effectiveness of different feature fusion strategies.
Our framework enhances the robot’s understanding of 3D spatial structures while completely eliminating the substantial costs associated with 3D point cloud acquisition.
Extensive experiments across various tasks validate that our framework can achieve performance comparable to 3D point cloud-based methods, without the actual point cloud data.
\end{abstract}


\section{Introduction}
In recent years, significant progress has been made in vision-based robotic manipulation strategies. Current approaches can be broadly classified into two categories. The first relies on 2D image inputs and employs reinforcement learning~\cite{{yarats2021mastering, lobbezoo2021reinforcement, chen2022towards, geng2023rlafford, an2024rgbmanip, hu2023imitation}} or imitation learning  ~\cite{fang2019survey,brohan2022rt,li2023manipllm, chi2023diffusion, zitkovich2023rt,  zhao2023learning,kim2024openvla, liu2024robomamba} to directly estimate the 3D end-effector pose.The second category utilizes 3D point clouds as input, leveraging their rich geometric information to build more accurate spatial perception~\cite{liu2022frame,shridhar2023perceiver, wang2024rise,jia2024lift3d}. Research ~\cite{zhu2024point} has shown that methods based on point cloud data generally outperform 2D image-based approaches in terms of policy performance and generalization.

Consequently, increasing attention has been directed toward explicitly extracting 3D feature representations for robotic manipulation. These approaches can be roughly divided into two groups. The first processes raw point cloud data by training 3D policy models from scratch~\cite{chen2023polarnet,ze20243d,zhang2024leveraging}. However, acquiring 3D data typically requires specialized sensors, and processing 3D or voxel-based features incurs high computational costs, limiting scalability and real-world deployment. The second group employs cross-modal transformation method—such as lifting 2D features into 3D space via pre-training or projecting 3D point clouds into multi-view 2D images processed by pre-trained 2D models~\cite{gervet2023act3d, ke20243d, xian2023chaineddiffuser, shridhar2022cliport,goyal2023rvt, goyal2024rvt, wang2024vihe, zhang2024sam}. However, these transformations often result in spatial information loss and lack end-to-end training capabilities.

Both approaches face challenges related to deployment cost, computational overhead, and adaptability. Therefore, we propose a central research question: \textbf{Can we develop a paradigm that enables robots to understand geometric spatial relationships solely from 2D images, without introducing any additional 3D input modalities?} Fortunately, 2D images inherently contain 3D structural cues~\cite{yang2024depthv2}, suggesting the possibility of building a 3D-perceiving policy based solely on image inputs. Inspired by this insight, we introduce a novel framework \textbf{NoReal3D},as shown in Figure~\ref{fig:pipeline}, that introduces the 3DStructureFormer,  a learnable 3D perception module module that to generate geometrically meaningful pseudo-point clouds from monocular images and fuses them with raw image features to enhance the robot's understanding of 3D spatial structures.

Specifically,  We first designed a pseudo point cloud generator that uses a monocular depth estimation model and camera intrinsic parameters to generate a pseudo point cloud to construct a 3D representation that preserves the image topology. We then design a pseudo-point cloud encoder that avoids traditional symmetric aggregation functions (e.g., max or min pooling), thereby preserving richer geometric details. Furthermore, we introduce a cross-modal fusion module that dynamically integrates 2D visual features with 3D geometric features, enhancing overall perception. To improve adaptability in real-world settings, we also propose an implicit 3D perception mechanism that can effectively estimate 3D structural information even when camera intrinsics are unknown or inaccurate.

We evaluate NoReal3D comprehensively on multiple simulated and real-world tasks , involving over 20 different grippers and dexterous manipulation scenarios. Experimental results show that our method outperforms existing 2D and 3D approaches \cite{majumdar2023we,nair2022r3m,he2016deep,banino2021pondernet,qi2017pointnet,ze20243d} on benchmarks such as ManiSkill and RLBench~\cite{gu2023maniskill2, james2020rlbench}. For example, on the RLBench benchmark, NoReal3D achieves an average success rate improvement of 10\% over prior 2D policy methods.In summary, this paper makes the following key contributions:
\begin{itemize}
\item We propose a novel 3D perception paradigm, \textbf{NoReal3D}, which integrates 3D representation mechanisms into existing 2D foundation models, enabling robots to perceive 3D spatial relationships using only 2D image inputs.  
\item We design the 3DStructureFormer and explore a cross-modal fusion strategy that effectively combines 2D visual features with 3D geometric information, significantly enhancing the spatial perception and manipulation capabilities of policy models.  
\item We validate the effectiveness of the proposed method across a wide range of simulated and real-world tasks, demonstrating its broad applicability and robust generalization in complex 3D manipulation scenarios.
\end{itemize}

\section{Related work}
\textbf{Manipulation Learning.}Manipulation Learning is a significant research direction at the intersection of robotics and artificial intelligence, dedicated to enabling robots to master complex physical manipulation skills such as grasping, assembly, and object manipulation through methods like reinforcement learning \cite{an2024rgbmanip,chen2022towards,geng2023rlafford}or deep learning\cite{radosavovic2023real,nair2022r3m,majumdar2023we,chen2024sugar}. This field aims to allow robots to autonomously learn from experience, optimizing performance through trial-and-error or by imitating human behaviors, thereby continuously improving the precision and environmental adaptability of their operations to achieve dexterous and robust autonomous manipulation.\\
\textbf{Robotic Representation Learning.}In embodied intelligence, observational inputs are primarily presented in the form of images and point clouds. Image data is low-cost and benefits from a vast number of available datasets. In this domain, image encoders aim to construct visual representations better suited to task requirements. R3M\cite{nair2022r3m} leverages temporal contrastive learning to capture sequential dynamics, while VC1\cite{majumdar2023we} and MVP\cite{radosavovic2023real} employ masked autoencoders, effectively mitigating domain shift issues. However, due to the lack of explicit 3D spatial structure, images face challenges in achieving precise spatial perception and geometric reasoning.Point cloud data, containing 3D spatial coordinates, provides rich geometric and appearance features. Unlike images, point cloud data is unordered. To address this disorder, existing point cloud encoding methods typically use symmetric functions for feature aggregation, which often leads to the loss of geometric details. Studies such as SUGAR\cite{chen2024sugar} and Point Cloud Matters\cite{zhu2024point} have compared image inputs with point cloud inputs, demonstrating that point cloud inputs generally enhance policy performance and generalization capabilities. However, acquiring high-quality point clouds relies on expensive specialized equipment, resulting in high costs that restrict the widespread application of point clouds in embodied intelligence.

\section{Methodology}  

The goal of robotic manipulation learning is to learn a policy $\pi(\cdot \mid \mathbf{o}_t, \tau)$ that maps visual observations $\mathbf{o}_t$ and a task specification $\tau$ to actions $\mathbf{a}_t$, with the aim of maximizing the expected cumulative binary reward for task completion. Using Behavioral Cloning (BC), the policy is trained on a dataset $\mathcal{D} = \{(X_i, \tau_i)\}_{i=1}^N$ consisting of $N$ pairs of successful demonstrations. Each demonstration $\tau_i$ comprises a sequence of $T$ timesteps, including visual observations $\{\mathbf{o}_t\}_{t=1}^T$ and corresponding actions $\{\mathbf{a}_t\}_{t=1}^T$ \cite{torabi2018behavioral}. The training objective minimizes a combined loss function composed of Mean Squared Error (MSE) on gripper position and rotation, and Binary Cross-Entropy (BCE) on the gripper open-close state:$\mathscr{L} = \frac{1}{|NT|} \sum_{\tau \in \mathcal{D}} \sum_{t=1}^{T} (\mathrm{MSE}(\hat{a}_t^{xyz}, a_t^{xyz}) + \mathrm{MSE}(\hat{a}_t^{q}, a_t^{q}) + \mathrm{BCE}(\hat{a}_t^{o}, a_t^{o}))$.

In this work, we focus on \textbf{robotic manipulation tasks driven solely by monocular RGB images}, where the observation $\mathbf{o}_t$ is a single image $\mathbf{I}_t\in\mathbb{R}^{H \times W \times 3}$. In fact, due to the lack of explicit 3D cues, conventional 2D vision-based policies often struggle to perceive accurate spatial relationships and geometric structures, limiting their generalization in complex manipulation scenarios. To address this, we propose \textbf{NoReal3D}, a novel learning paradigm that integrates both 2D vision with Pseudo 3D representations. Specifically, we introduce the \textbf{3DStructureFormer}, a learnable 3D perception module that takes a monocular image $\mathbf{I}_t$ as input and generates geometrically meaningful pseudo-point cloud features $\mathbf{F}^{\text{3d}}_t\in\mathbb{R}^{H' \times W' \times C}$ via Pseudo Point Cloud Generator and Pseudo Point Cloud Encoder. This pseudo-point cloud features is then fused with 2D visual features to produce a 3D-enhanced representation $\mathbf{F}^{\text{fused}}_t\in\mathbb{R}^{H' \times W' \times C}$. The strategy can be \textbf{applied by the existing 2D foundation models} (e.g., ViT \cite{dosovitskiy2020image} or ResNet \cite{he2016deep}) in a plug-and-play manner, enhancing their spatial awareness without modifying the backbone architecture.To the best of our knowledge, this work represents the first effort to enable 2D-to-3D lifting within a robotic manipulation framework without relying on auxiliary input information.
\subsection{3DStructureFormer}
\label{subsec:3dformer}

The term \textit{3DStructureFormer} refers to the core pipeline of NoReal3D that transforms 2D images into 3D-structured representations. This module does not operate on real point cloud data but instead constructs a \textit{pseudo-point cloud} from monocular inputs, preserving both appearance and geometric topology.
3DStructureFormer consists of two main submodules: the Pseudo Point Cloud Generator and the Pseudo Point Cloud Encoder.

\subsubsection{Pseudo Point Cloud Generator}
\label{sssec:generator}

\maketitle

In computer vision and 3D reconstruction, given the camera intrinsic parameters and a depth map, we can convert the depth value of each pixel into a 3D point to generate a point cloud.The corresponding 3D coordinates $ (X_w, Y_w, Z_w) $ in the camera coordinate system can be computed as:

$$
\left\{
\begin{aligned}
X_w &= d_{\text{truth}} \cdot \frac{u - c_x}{f_x} \\
Y_w &= d_{\text{truth}} \cdot \frac{v - c_y}{f_y} \\
Z_w &= d_{\text{truth}}
\end{aligned}
\right.
$$

This formulation is based on the pinhole camera model, which back-projects 2D image coordinates scaled by depth into 3D space, enabling the construction of dense point clouds.

Considering that in our scene, there are only RGB images $\mathbf{I}$  and no absolute depth map information, we can employ depth estimation models $\mathcal{M}_{\text{depth}}$  to predict \textbf{relative depth} $ d_{\text{pred}} $. Ideally, the predicted depth $ d_{\text{pred}} $ is affinely related to the inverse of the true depth (i.e., disparity):

$$
d_{\text{pred}} = s \cdot \frac{1}{d_{\text{gt}}} + t
$$
where $ d_{\text{gt}} $ denotes the true depth, $ s $ is a scale factor, and $ t $ is a shift parameter. While the scale $ s $ does not affect the overall geometric structure of the point cloud (only changing its global scale), the shift $ t $ distorts the inverse relationship and leads to shape deformation in the reconstructed 3D points.

Notably, \textbf{affine transformations (scale and shift) do not affect the normalized depth map}. Specifically, normalization is invariant under affine transformations:
$$
\frac{d_{\text{pred}} - \min(d_{\text{pred}})}{\max(d_{\text{pred}}) - \min(d_{\text{pred}})} = \frac{\frac{1}{d_{\text{gt}}} - \min\left(\frac{1}{d_{\text{gt}}}\right)}{\max\left(\frac{1}{d_{\text{gt}}}\right) - \min\left(\frac{1}{d_{\text{gt}}}\right)}
$$

Therefore, to recover a geometrically plausible point cloud from relative depth, we first normalize $ d_{\text{pred}} $ into the $[0, 1]$ range:
$$
d_{\text{nor}} = \frac{d_{\text{pred}} - \min(d_{\text{pred}})}{\max(d_{\text{pred}}) - \min(d_{\text{pred}})}
$$

Since depth prediction models typically output values inversely correlated with depth (smaller values for distant regions), we invert the normalized depth to ensure positive correlation with true depth:
$$
d_r = 1 - d_{\text{nor}}
$$

Finally, we substitute $ d_r $ into the pinhole camera model to generate a pseudo-point cloud:


\begin{align}
\label{eq:backproj}
\left\{
\begin{aligned}
X_{\text{Pseudo}} &= d_r \cdot \frac{u - c_x}{f_x} \\
Y_{\text{Pseudo}} &= d_r \cdot \frac{v - c_y}{f_y} \\
Z_{\text{Pseudo}} &= d_r
\end{aligned}
\right.
\end{align}

This approach enables the reconstruction of point clouds with reasonable geometry and relative proportions, even without absolute depth supervision.

Specifically, given an input image $\mathbf{I} \in \mathbb{R}^{H \times W \times 3}$, we first estimate its dense depth map 
$
\mathbf{d}_{\text{pred}} = \mathcal{M}_{\text{depth}}(\mathbf{I}) \in \mathbb{R}^{H \times W \times 3}
$
using a pre-trained depth estimation model $\mathcal{M}_{\text{depth}}$, implemented as a lightweight encoder-decoder network. This model is optimized end-to-end during training without requiring ground-truth depth supervision. Subsequently, using the camera intrinsic matrix $\mathbf{K}$, each pixel $(u, v)$ is back-projected into 3D space according to Equation~\eqref{eq:backproj} to generate the pseudo-point cloud  $\mathcal{P} = \{\mathbf{P}_{uv}\} \in \mathbb{R}^{H \times W \times 3}$.The entire process can be summarized as:
\begin{equation}
    \mathcal{P}= \mathcal{G}_{\text{3d}}(\mathbf{I})
\end{equation}


The resulting pseudo-point cloud  preserves the 2D topological structure of the original image (i.e., neighboring pixels remain locally continuous in 3D) and encodes preliminary geometric layout. Notably, the pseudo-point cloud is organized in a regular grid, allowing it to be treated as a voxelized representation and processed directly by standard vision networks.

\subsubsection{Pseudo Point Cloud Encoder}
Traditional point cloud encoders, such as those based on PointNet or its hierarchical variants, typically operate on unstructured 3D point sets and rely on symmetric aggregation functions (e.g., max or min pooling) to achieve permutation invariance. While effective for general 3D recognition, these designs inherently discard precise spatial relationships and local topology, limiting their ability to model fine-grained geometric structures critical in robotic perception tasks.

In contrast, our Pseudo Point Cloud Encoder explicitly preserves and leverages topological structure by maintaining the 2D grid organization of the input. Instead of processing raw 3D points directly, we first project the generated pseudo-point cloud $\mathcal{P}$ into a structured 3-channel coordinate map:
\begin{equation}
    \mathcal{P}_{\text{3d}} = [ \mathbf{X}, \mathbf{Y}, \mathbf{Z}] \in \mathbb{R}^{H \times W \times 3},
\end{equation}
where each pixel encodes the $(x, y, z)$ coordinates of the corresponding 3D point. This representation naturally retains the spatial continuity and neighborhood relationships present in the original 2D sensor layout (e.g., from RGB cameras), effectively embedding 3D geometry into a topology-aware 2D domain.

The structured tensor $\mathbf{I}_{\text{3d}}$ is then fed into a slightly modified 2D vision backbone $\mathcal{E}_{\text{3d}}$, such as ResNet or Vision Transformer, to produce geometry-enriched feature maps $\mathbf{F}^{\text{3d}} \in \mathbb{R}^{H' \times W' \times C}$. By avoiding hand-crafted point cloud operators and symmetric pooling layers, our encoder allows convolutional filters or self-attention mechanisms to naturally capture both local geometric patterns and long-range spatial dependencies within a consistent topological framework. This is particularly advantageous in robotic scenarios where spatial coherence and precise localization are crucial.

Furthermore, our design enables seamless integration with off-the-shelf 2D foundation models. The encoder can be frozen for efficient transfer, fine-tuned, or trained end-to-end depending on task requirements and data availability. The final 3D-aware representation is obtained as:
\begin{equation}
    \mathbf{F}^{\text{3d}} = \mathcal{E}_{\text{3d}}(\mathcal{P}),
\end{equation}
which is both computationally efficient and highly expressive. In summary, by transforming unstructured point clouds into structured coordinate maps, our Pseudo Point Cloud Encoder introduces explicit topological modeling into 2D backbones—overcoming a key limitation of conventional point-based encoders—while eliminating the need for specialized and often inefficient point cloud operations.

\subsection{Cross-Modal Fusion Module}

To integrate the appearance-rich 2D features $\mathbf{F}^{\text{2d}}$ from $\mathcal{E}_{\text{2d}}$ with the geometry-aware 3D features $\mathbf{F}^{\text{3d}}$ from $\mathcal{E}_{\text{3d}}$, we systematically explore several fusion strategies, including addition, concatenation, cross-attention, and self-attention mechanisms. In addition fusion, the 2D and 3D features are directly summed: $\mathbf{F}^{\text{fused}} = \mathbf{F}^{\text{2d}} + \mathbf{F}^{\text{3d}}$, enabling a simple and efficient combination. Alternatively, concatenation fusion preserves the integrity of both feature sets by concatenating them along the channel dimension, $\mathbf{F}^{\text{fused}} = [\mathbf{F}^{\text{2d}}, \mathbf{F}^{\text{3d}}]$, followed by a $1\times1$ convolution to reduce dimensionality. To enable more sophisticated interactions, we also investigate cross-attention fusion, where 2D features serve as queries and 3D features as keys and values, formulated as $\mathbf{F}^{\text{fused}} = \text{MHCA}(\mathbf{F}^{\text{2d}}, \mathbf{F}^{\text{3d}}, \mathbf{F}^{\text{3d}})$, allowing 2D features to selectively attend to relevant geometric cues in 3D space. Lastly, self-attention fusion employs a Transformer encoder layer on the concatenated features: $\mathbf{F}^{\text{fused}} = \text{SA}([\mathbf{F}^{\text{2d}}, \mathbf{F}^{\text{3d}}])$, aiming to model complex interdependencies.

Among these, ablation studies reveal that addition fusion consistently achieves the best performance, offering effective feature integration with minimal computational overhead. Detailed experimental analyses of these fusion mechanisms are provided in Feature Fusion Experiments .

 We propose NoReal3D, a novel robotic manipulation learning framework that enhances 2D vision-based policies with pseudo-3D geometric understanding from monocular RGB images. Our method is plug-and-play compatible with existing 2D foundation models, marking the first effort to enable 2D-to-3D lifting in robotic manipulation without auxiliary inputs.

\section{Experiments}
\begin{table*}[t]
\setlength{\tabcolsep}{3pt}
\caption{Experimental results of different models combined with the ACT strategy on RLBench. }
\label{tab:RLBench_ACT}
\centering

\begin{tabular}{l| *{8}{c}| *{2}{c}} 

\hline 
\multirow{3}{*}{\textbf{Taskname}} & 
\multicolumn{8}{c|}{\textbf{2D Model + ACT}} & 
\multicolumn{2}{c}{\textbf{3D Model + ACT}} \\
\cline{2-11} 
& \multirow{2}{*}{\textbf{vc1}} & \textbf{vc1} & \multirow{2}{*}{\textbf{r3m}} & \textbf{r3m} & 
\multirow{2}{*}{\textbf{vit}} & \textbf{vit} & \multirow{2}{*}{\textbf{resnet}} & \textbf{resnet} & 
\multirow{2}{*}{\textbf{PointNet}} & \multirow{2}{*}{\textbf{PonderV2}} \\
 & & \textbf{Ours}& &\textbf{Ours}& &\textbf{Ours}& &\textbf{Ours}& & \\
\hline 
close jar& 0.08 & \textbf{\underline{0.32}} & 0.00 & 0.00 & 0.00 & \underline{0.04} & 0.00 & 0.00 & 0.08 & 0.24\\
meat off grill& 0.28 & 0.28 & 0.00 & 0.00 & 0.00 & 0.00 & 0.00 & 0.00 & 0.72 & \textbf{0.76} \\
open drawer& 0.16 & \underline{0.40} & 0.00 & \underline{0.12} & 0.04 & \underline{0.28} & 0.08 & \underline{0.36} & 0.04 & \textbf{0.48} \\
place wine& 0.04 & \underline{0.12} & 0.00 & 0.00 & 0.00 & 0.00 & 0.00 & 0.00 & 0.00 & \textbf{0.36} \\
push buttons& 0.36 & \underline{0.44} & 0.04& \underline{0.12} & 0.00 & \underline{0.20} & 0.00 & \underline{0.16} & \textbf{0.68} & 0.52 \\
put money& 0.52 & \textbf{\underline{0.56}} & 0.16 & 0.08 & 0.00 & \underline{0.44} & 0.04 & 0.04 & 0.04 & 0.28 \\
reach and drag& 0.20 & 0.20 & 0.16 & 0.08 & 0.00 & \underline{0.64} & 0.00 & \underline{0.04} & \textbf{0.76} & 0.28 \\
sweep to& 0.72 & \textbf{\underline{0.96}} & 0.44& \underline{0.92} & 0.00 & \underline{0.40} & 0.00 & \underline{0.36} & \textbf{0.96} & 0.92 \\
turn tap& 0.00 & 0.00 & 0.00 & 0.00 & 0.00 & 0.00 & 0.00 & 0.00 & 0.04 & \textbf{0.16} \\
\hline 
\textbf{Mean S.R.} &0.26&\underline{0.36}&0.09&\underline{0.15}&0.00&\underline{0.22}&0.01&\underline{0.11} &0.37&\textbf{0.44}\\
\hline
\end{tabular}
\end{table*}

\begin{table*}[t]
\caption{Experimental results of different models combined with the Diffusion Policy strategy on RLBench. }
\setlength{\tabcolsep}{3pt}
\label{tab:RLBench_DP}
\centering

\begin{tabular}{l| *{8}{c}| *{2}{c} | c} 
\hline 
\multirow{3}{*}{\textbf{Taskname}} & 
\multicolumn{8}{c |}{\textbf{2D Model + DP}}  & 
\multicolumn{2}{c |}{\textbf{3D Model + DP}} & {\textbf{3D Policy}}\\

\cline{2-12} 
& \multirow{2}{*}{\textbf{vc1}} & \textbf{vc1} & \multirow{2}{*}{\textbf{r3m}} & \textbf{r3m} & 
\multirow{2}{*}{\textbf{vit}} & \textbf{vit} & \multirow{2}{*}{\textbf{resnet}} & \textbf{resnet} & 
\multirow{2}{*}{\textbf{PointNet}} & \multirow{2}{*}{\textbf{PonderV2}} & \multirow{2}{*}{\textbf{DP3}}
\\
 & & \textbf{Ours}& &\textbf{Ours}& &\textbf{Ours}& &\textbf{Ours}& & \\
\hline 
close jar& 0.00 & 0.00 & 0.00 & 0.00 & 0.00 & \textbf{\underline{0.04}} & 0.00 & 0.00 & 0.00 & 0.00 & 0.00\\
meat off grill& 0.00 & 0.00 & 0.00 & 0.00 & 0.00 & 0.00 & 0.32 & \textbf{\underline{0.40}} & 0.00 & 0.00 & 0.00\\
open drawer& 0.20 & \underline{0.32} & 0.00 & \underline{0.08} & 0.08 & \underline{0.16} & 0.56 & \textbf{\underline{0.68}} & 0.08 & 0.16 & 0.04\\
place wine& 0.00 & \textbf{\underline{0.04}} & 0.00 & 0.00 & 0.00 & 0.00 & 0.00 & 0.00 & 0.00 & 0.00 & 0.00\\
push buttons& 0.08 & \textbf{\underline{0.12}} & 0.00& \underline{0.08} & 0.00 & \underline{0.00} & 0.04 & \underline{0.08} & 0.00 & 0.04 & 0.12\\
put money& \underline{0.24} & 0.16 & 0.04 & \underline{0.12} & 0.16 & \textbf{\underline{0.52}} & 0.16 & \underline{0.20} & 0.04 & 0.04 & 0.00\\
reach and drag& 0.36 & \textbf{\underline{0.76}} & 0.00 & 0.00 & 0.08 & \underline{0.12} & \underline{0.48} & 0.40 & 0.16 & 0.04 & 0.04\\
sweep to& \underline{0.36} & 0.28 & 0.00& \underline{0.44} & 0.72 & \underline{0.84} & \textbf{0.96} & \textbf{0.96} & 0.28 & 0.12 & 0.00\\
turn tap& 0.12 & \textbf{\underline{0.16}} & 0.00 & 0.00 & 0.12 & 0.12 & \underline{0.12} & 0.08 & 0.04 & 0.00 & 0.00\\
\hline 
\textbf{Mean S.R.} &0.15&\underline{0.20}&0.00&\underline{0.08}&0.13&\underline{0.21}&0.29&\textbf{\underline{{0.31}}} &0.07&0.04 & 0.02\\
\hline
\end{tabular}
\end{table*}




To validate the effectiveness and generalization capability of NoReal3D, we conducted systematic experiments combining various encoders and action generation strategies in two simulated environments and a real-world scenario. We also investigated the impact of different feature fusion strategies on the results and provided our analysis.To verify the effectiveness of the relative depth processing, we also conducted visualization experiments on pseudo point clouds.Due to space limitations, the experimental results of the implicit branch without camera intrinsics will be presented in the Appendix~A.
\subsection{Experiment Settings}
\label{sec:sim-exp}
\textbf{Benchmarks and dataset.} We evaluate on over 20 tasks selected from two widely used manipulation simulation benchmarks: \textbf{RLBench}\cite{james2020rlbench}, based on the CoppeliaSim physics engine, and \textbf{ManiSkill2}\cite{gu2023maniskill2}, based on the SAPIEN physics engine\cite{Xiang_2020_SAPIEN}.On RLBench, we use the pre-generated demonstration data for 18 tasks provided by PerAct, with each task containing 100 demonstrations for training and 25 demonstrations for testing and validation. On ManiSkill2, we use the official demonstration dataset: soft-body manipulation tasks contain 1,000 demonstrations, while rigid-body manipulation tasks contain 200 demonstration samples.In the real-world experiments, for each task, we collected 50 demonstration trajectories from different spatial locations at a sampling frequency of 30 fps.Detailed task descriptions are provided in Appendix~B.\\
\textbf{Backbone.}We conducted experiments on two state-of-the-art action generation strategies, ACT \cite{fu2024mobile} and Diffusion Policy(DP)\cite{chi2023diffusion}, as our action generation frameworks. The core idea of ACT is to utilize a Variational Autoencoder (VAE) to learn and encode action sequences from expert demonstration data, generating coherent action sequences through imitation learning. Diffusion Policy, on the other hand, is based on diffusion models and formulates action generation as a step-by-step denoising process, iteratively recovering high-quality and diverse actions from noise.\\
\textbf{Baselines.}To verify the applicability of our method across different feature extraction architectures, we select four representative 2D image encoders as baseline models: VC1\cite{dosovitskiy2020image}, R3M\cite{nair2022r3m}, ViT-B\cite{dosovitskiy2020image}, and ResNet-50\cite{he2016deep}. These encoders cover two mainstream network architectures (Transformers and ResNet) including general-purpose visual encoders (ViT-B, ResNet-50) and specialized encoders pre-trained on embodied intelligence data (VC1, R3M). Additionally, to extend the scope of comparison, we also introduced two 3D point cloud encoders—PointNet\cite{qi2017pointnet} and PonderV2\cite{banino2021pondernet}—and included DP3\cite{ze20243d}, the current state-of-the-art 3D perception policy method, in the comparative experiments as one of the 3D model baselines, thereby comprehensively evaluating the performance of the proposed method across different architectural types.\\
\textbf{Training Details.}To ensure a fair comparison, all models are trained and evaluated under identical configurations within the same virtual environment. During training, the AdamW\cite{kingma2014adam} optimizer is employed with a weight decay of 0.05 and a learning rate of $1 \times 10^{-5}$. For the RLBench experiments, models are trained for 3,000 epochs; for the ManiSkill experiments, models are trained for 1,200 epochs. Performance is evaluated using the final checkpoint obtained from training. Each task's checkpoint is independently evaluated 25 times, and the average success rate is reported as the final result. All models are trained and evaluated on a single RTX 4090 GPU. 

\begin{table*}[t]

\caption{Results of Different Feature Fusion Strategies. VC1 encoder is used as the 2D encoder. Best performances are highlighted in bold.}
\label{tab:feature_fusion}
\centering

\begin{tabular}{l| *{8}{c}| *{2}{c}} 
\hline 
\multirow{2}{*}{\textbf{Taskname}}
& \multirow{2}{*}{\textbf{plus}} & \textbf{plus} & \multirow{2}{*}{\textbf{contact}} & \textbf{contact} & 
 \textbf{cross}  \\
 & & \textbf{Attention}& &\textbf{Attention}&\textbf{Attention}\\
\hline 
close jar& \textbf{0.32} & 0.08 & 0.16 & 0.12 & 0.12 \\
meat off grill& 0.28 & 0.00 & \textbf{0.20} & 0.12 & 0.16 \\
open drawer& \textbf{0.40} & 0.24 & 0.24 & 0.16 & 0.24 \\
place wine& \textbf{0.12} & 0.04 & 0.00 & 0.00 & 0.00 \\
push buttons& 0.44 & 0.44 & \textbf{0.52} & 0.40 & 0.40 \\
put money& \textbf{0.56} & 0.48 & \textbf{0.56} & 0.44 & 0.48 \\
reach and drag& 0.20 & 0.12 & \textbf{0.36} & 0.20 & 0.04 \\
sweep to& \textbf{0.96} & 0.56 & 0.80 & 0.80 & 0.80 \\
turn tap& 0.00 & 0.00 & \textbf{0.04} & \textbf{0.04} & 0.00 \\
\hline 
\textbf{Mean S.R.} &\textbf{0.36}&0.22&0.32&0.25&0.25\\
\hline
\end{tabular}
\end{table*}
\begin{figure*}
    \centering 
    \subfigure[Origin Image.]{%
        \resizebox*{5cm}{!}{\includegraphics{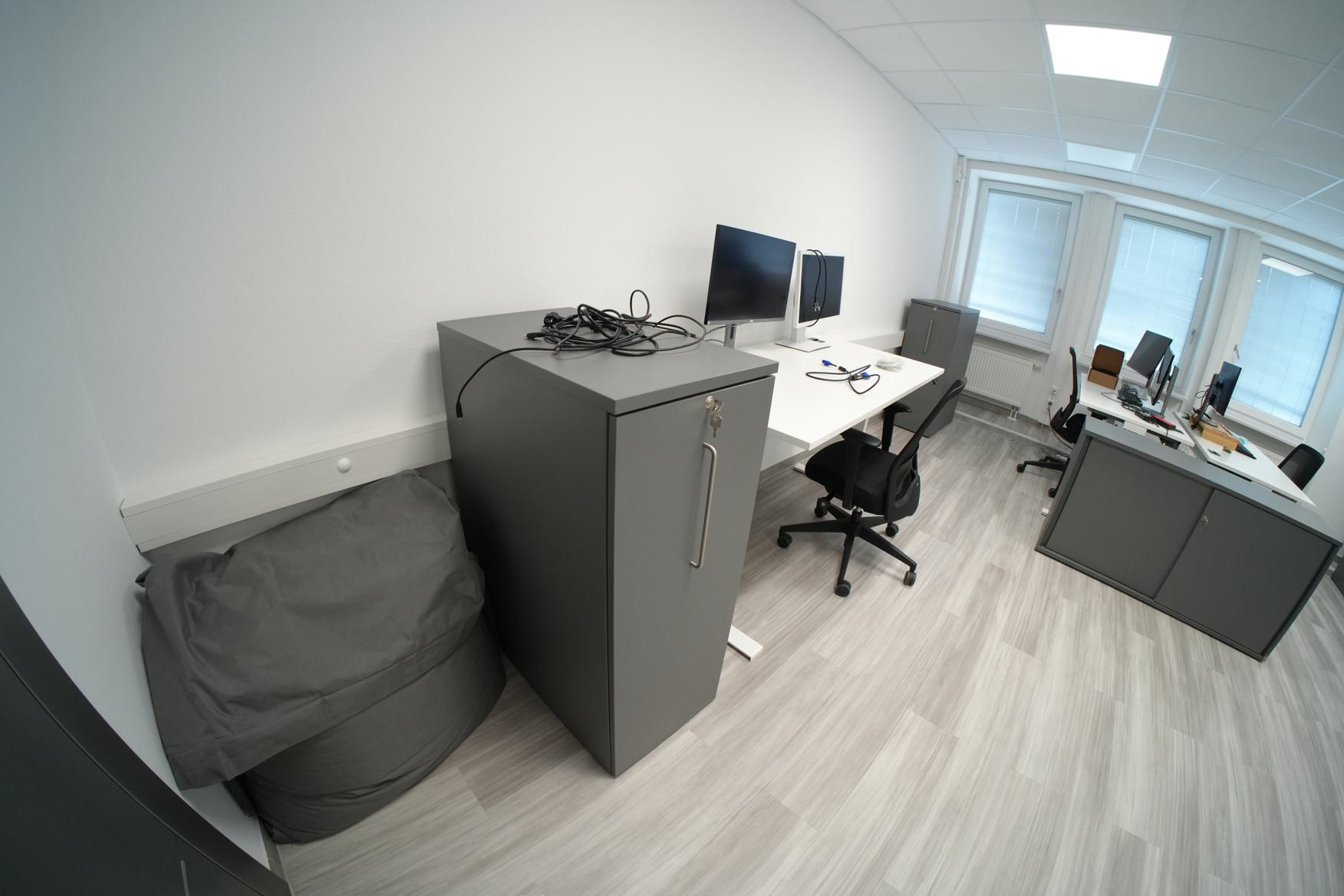}}}\hspace{5pt}
    \subfigure[
Taking the reciprocal of relative depth.]{%
        \resizebox*{6cm}{!}{\includegraphics{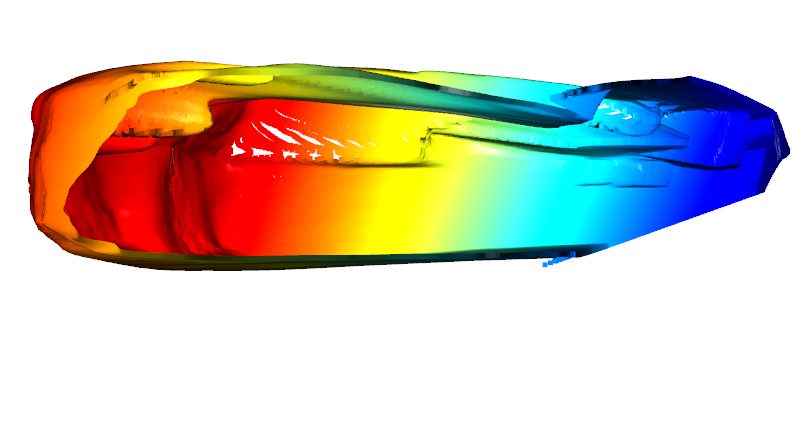}}}\hspace{5pt}
    \subfigure[NoReal3D.]{%
        \resizebox*{5cm}{!}{\includegraphics{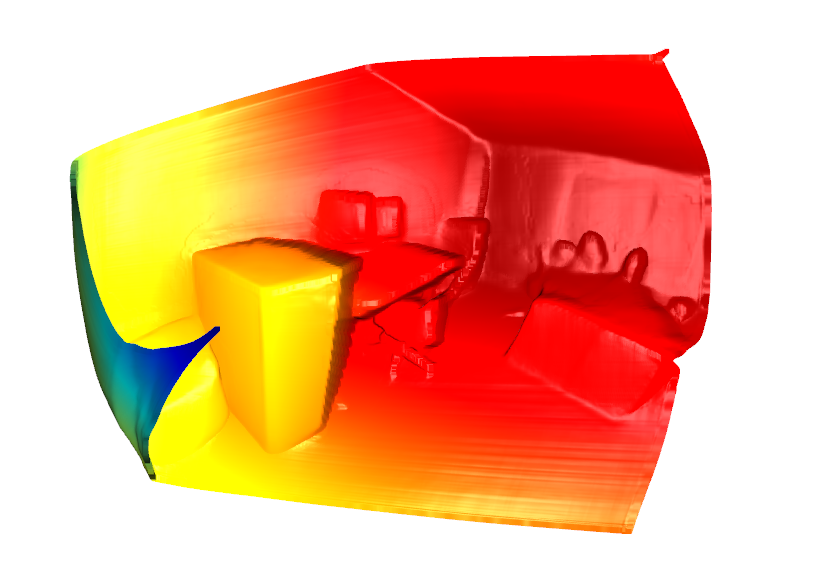}}}
    \caption{Pseudo-point cloud visualization experiment. (a) is an image taken from ScanNet++; (b) is the pseudo-point cloud generated by taking the reciprocal of the relative depth and using camera intrinsic parameters; (c) is the pseudo-point cloud generated by NoReal3D after normalization and other processing.} \label{fig:Pseudo_point}
\end{figure*}
\subsection{Simulation Experiment}
In presenting the results, for each 2D model and its corresponding version enhanced by our method, the better-performing one is \underline{underlined}; the best overall result among all 2D and 3D models for each task is \textbf{highlighted in bold} to facilitate clear comparison and analysis.As shown in Table~\ref{tab:RLBench_ACT} and Table~\ref{tab:RLBench_DP}, NoReal3D significantly improves the performance of specialized encoders pre-trained on embodied intelligence data. For the VC1 encoder, under the ACT policy, the average task success rate increases by 0.10, representing a relative improvement of 38\%; under the DP policy, the average task success rate increases by 0.05, a relative improvement of 33\%. For the R3M encoder, under the ACT policy, the average task success rate increases by 0.05, a relative improvement as high as 55\%; under the DP policy, the success rate improves from nearly zero to 0.08, demonstrating particularly significant enhancement. Furthermore, NoReal3D also exhibits strong adaptability with general-purpose encoders. Under the ACT policy, the previously nearly ineffective ViT and ResNet models achieve success rates of 0.22 and 0.10, respectively, marking a transition from almost useless to practically applicable. These results fully demonstrate NoReal3D's excellent compatibility and strong generalization capabilities across different 2D encoders and various action generation strategies. Compared to 3D baseline methods, NoReal3D combined with the state-of-the-art 2D encoder VC1 approaches the performance of the 3D baselines on many tasks under the ACT policy, while under the DP policy, it surpasses the 3D baselines on all tasks, demonstrating superior practicality and competitiveness.Due to space limitations, the experimental results and quantitative analysis of ManiSkill2 are provided in the Appendix~C.

\subsection{Real-World Experiments}
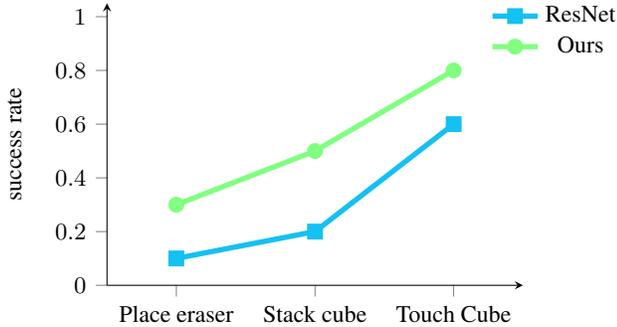
\begin{figure}[ht]
\centering
\begin{tikzpicture}
\begin{axis}[
    width=0.4\textwidth,
    height=0.3\textwidth,
    ymin=0, ymax=1.05, 
    xmin=0, xmax=3,
    enlargelimits=false,
    ytick={0,0.2,0.4,0.6,0.8,1.0},
    ylabel={success rate},
    xtick={0.5,1.5,2.5},
    xticklabels={Place eraser, Stack cube, Touch Cube},
    legend style={
        at={(0.9,0.9)},
        anchor=west,
        draw=none,
        fill=none,
        font=\small
    },
    axis line style={-}, 
    axis x line=bottom,
    axis y line=left,
    axis on top,
    tick align=outside,
    tick label style={font=\small},
    label style={font=\small}
]
\addplot[
    color=cyan!70,
    mark=square*,
    line width=2pt
]
coordinates {(0.5,0.1) (1.5,0.2) (2.5,0.6)};

\addplot[
    color=green!50,
    mark=*,
    line width=2pt
]
coordinates {(0.5,0.3) (1.5,0.5) (2.5,0.8)};

\legend{ResNet, Ours}

\end{axis}

\end{tikzpicture}
\caption{Success rates comparison between ResNet and Ours methods on three tasks.}
\label{fig:successRateComp}
\end{figure}

\textbf{Task Design.} We conduct real-world experiments using a WOWROBO single-arm manipulator (Koch type), acquiring visual inputs from two external cameras—a Hikvision DS-E22S camera and a GESOBYTE C301 camera (total cost under 40 USD)—from different viewpoints to construct dual-view observations. The experiments involve three manipulation tasks: (1) stacking a cubic block onto a rectangular block; (2) placing an eraser into a designated container;  and (3) touching cubic block. All policies are trained independently from scratch, and after training, each is evaluated independently 10 times at unseen locations. A ResNet architecture is used for the 2D visual encoder. Camera intrinsic parameters are estimated based on field of view and resolution, ignoring lens distortion. More experimental setup and implementation details are provided in the Appendix~D.\\
\textbf{Quantitative results.}Compared to baseline methods using only a 2D encoder, our approach achieves significant success rate improvements across all three tasks. This improvement validates that our method can effectively enhance the 2D encoder's understanding of 3D spatial structures. Even without using expensive 3D sensors or pre-scanned point clouds, our approach significantly improves the policy's spatial perception and generalization performance by implicitly modeling multi-view geometric information.
\subsection*{Feature Fusion Experiments}

\textbf{Results and Discussion.}As shown in Table~\ref{tab:feature_fusion}, for most tasks, simple element-wise addition of features performs better than channel-wise concatenation, and both methods outperform subsequent approaches that incorporate attention mechanisms. This phenomenon may be attributed to image features primarily carrying semantic information, while pseudo point cloud features focus more on spatial information expression. Given the one-to-one correspondence between pseudo point cloud features and image features, directly adding the two can simultaneously preserve semantic and spatial information at each resolution level, whereas concatenation might disrupt this precise positional correspondence. Moreover, attention mechanisms aim to enhance the representation of useful information by adjusting feature weights. However, for pseudo point clouds generated from images, which essentially approximate rather than precisely replicate actual point cloud positions, applying attention mechanisms after addition or concatenation could lead the model to over-focus on these approximation errors, thereby weakening overall performance.
\subsection*{Pseudo Point Cloud Visualization Experiments}
\textbf{Experimental Design.} We employ Depth Anything as the relative depth estimation model and select an image from the ScanNet++\cite{yeshwanth2023scannet++} dataset to demonstrate the visualization of pseudo point clouds. The experiment compares two generation methods: one generates pseudo point clouds by directly taking the inverse of the relative depth map, while the other produces point clouds after applying normalization and other processing steps within the NoReal3D framework. This comparison visually illustrates the impact of different processing strategies on point cloud quality.\\
\textbf{Experimental Analysis.}As shown in Figure~\ref{fig:Pseudo_point},due to the presence of an unknown offset, directly inverting the relative depth map results in significant positional misalignment in the reconstructed point cloud. In contrast, NoReal3D effectively eliminates the impact of this offset through normalization and other optimization techniques, producing a detailed point cloud that closely aligns with the real scene. Fine details present in the original image, such as chair legs and other small structures, are well preserved in the reconstructed result.

\section{Conclusion}

In this paper, we introduce NoReal3D, a novel framework that enables robotic manipulation from monocular RGB images by fusing 2D vision with pseudo 3D representations. First, we propose the 3DStructureFormer, a learnable module that generates geometrically meaningful pseudo-point cloud features from single 2D images, enhancing spatial and structural perception. Second, we design a cross-modal fusion strategy that seamlessly integrates 3D geometric cues with 2D visual features in a plug-and-play manner, boosting the performance of existing 2D foundation models without modifying their architectures. NoReal3D achieves state-of-the-art results on benchmarks such as ManiSkill and RLBench, reaching performance on par with state-of-the-art 3D policy methods. It also demonstrates strong capabilities in real-world tasks.

\bibliography{aaai2026}

\makeatletter
\@ifundefined{isChecklistMainFile}{
  \newif\ifreproStandalone
  \reproStandalonetrue
}{
  \newif\ifreproStandalone
  \reproStandalonefalse
}
\makeatother

\ifreproStandalone
\documentclass[letterpaper]{article}
\usepackage[submission]{aaai2026}
\setlength{\pdfpagewidth}{8.5in}
\setlength{\pdfpageheight}{11in}
\usepackage{times}
\usepackage{helvet}
\usepackage{courier}
\usepackage{xcolor}
\frenchspacing

\begin{document}
\fi
\setlength{\leftmargini}{20pt}
\makeatletter\def\@listi{\leftmargin\leftmargini \topsep .5em \parsep .5em \itemsep .5em}
\def\@listii{\leftmargin\leftmarginii \labelwidth\leftmarginii \advance\labelwidth-\labelsep \topsep .4em \parsep .4em \itemsep .4em}
\def\@listiii{\leftmargin\leftmarginiii \labelwidth\leftmarginiii \advance\labelwidth-\labelsep \topsep .4em \parsep .4em \itemsep .4em}\makeatother

\setcounter{secnumdepth}{0}
\renewcommand\thesubsection{\arabic{subsection}}
\renewcommand\labelenumi{\thesubsection.\arabic{enumi}}

\newcounter{checksubsection}
\newcounter{checkitem}[checksubsection]

\newcommand{\checksubsection}[1]{%
  \refstepcounter{checksubsection}%
  \paragraph{\arabic{checksubsection}. #1}%
  \setcounter{checkitem}{0}%
}

\newcommand{\checkitem}{%
  \refstepcounter{checkitem}%
  \item[\arabic{checksubsection}.\arabic{checkitem}.]%
}
\newcommand{\question}[2]{\normalcolor\checkitem #1 #2 \color{blue}}
\newcommand{\ifyespoints}[1]{\makebox[0pt][l]{\hspace{-15pt}\normalcolor #1}}

\section*{Reproducibility Checklist}

\vspace{1em}
\hrule
\vspace{1em}

\textbf{Instructions for Authors:}

This document outlines key aspects for assessing reproducibility. Please provide your input by editing this \texttt{.tex} file directly.

For each question (that applies), replace the ``Type your response here'' text with your answer.

\vspace{1em}
\noindent
\textbf{Example:} If a question appears as
\begin{center}
\noindent
\begin{minipage}{.9\linewidth}
\ttfamily\raggedright
\string\question \{Proofs of all novel claims are included\} \{(yes/partial/no)\} \\
Type your response here
\end{minipage}
\end{center}
you would change it to:
\begin{center}
\noindent
\begin{minipage}{.9\linewidth}
\ttfamily\raggedright
\string\question \{Proofs of all novel claims are included\} \{(yes/partial/no)\} \\
yes
\end{minipage}
\end{center}
Please make sure to:
\begin{itemize}\setlength{\itemsep}{.1em}
\item Replace ONLY the ``Type your response here'' text and nothing else.
\item Use one of the options listed for that question (e.g., \textbf{yes}, \textbf{no}, \textbf{partial}, or \textbf{NA}).
\item \textbf{Not} modify any other part of the \texttt{\string\question} command or any other lines in this document.\\
\end{itemize}

You can \texttt{\string\input} this .tex file right before \texttt{\string\end\{document\}} of your main file or compile it as a stand-alone document. Check the instructions on your conference's website to see if you will be asked to provide this checklist with your paper or separately.

\vspace{1em}
\hrule
\vspace{1em}


\checksubsection{General Paper Structure}
\begin{itemize}

\question{Includes a conceptual outline and/or pseudocode description of AI methods introduced}{(yes/partial/no/NA)}
yes

\question{Clearly delineates statements that are opinions, hypothesis, and speculation from objective facts and results}{(yes/no)}
yes

\question{Provides well-marked pedagogical references for less-familiar readers to gain background necessary to replicate the paper}{(yes/no)}
yes

\end{itemize}
\checksubsection{Theoretical Contributions}
\begin{itemize}

\question{Does this paper make theoretical contributions?}{(yes/no)}
no

	\ifyespoints{\vspace{1.2em}If yes, please address the following points:}
        \begin{itemize}
	
	\question{All assumptions and restrictions are stated clearly and formally}{(yes/partial/no)}
	Type your response here

	\question{All novel claims are stated formally (e.g., in theorem statements)}{(yes/partial/no)}
	Type your response here

	\question{Proofs of all novel claims are included}{(yes/partial/no)}
	Type your response here

	\question{Proof sketches or intuitions are given for complex and/or novel results}{(yes/partial/no)}
	Type your response here

	\question{Appropriate citations to theoretical tools used are given}{(yes/partial/no)}
	Type your response here

	\question{All theoretical claims are demonstrated empirically to hold}{(yes/partial/no/NA)}
	Type your response here

	\question{All experimental code used to eliminate or disprove claims is included}{(yes/no/NA)}
	Type your response here
	
	\end{itemize}
\end{itemize}

\checksubsection{Dataset Usage}
\begin{itemize}

\question{Does this paper rely on one or more datasets?}{(yes/no)}
yes

\ifyespoints{If yes, please address the following points:}
\begin{itemize}

	\question{A motivation is given for why the experiments are conducted on the selected datasets}{(yes/partial/no/NA)}
	yes

	\question{All novel datasets introduced in this paper are included in a data appendix}{(yes/partial/no/NA)}
	yes

	\question{All novel datasets introduced in this paper will be made publicly available upon publication of the paper with a license that allows free usage for research purposes}{(yes/partial/no/NA)}
	yes

	\question{All datasets drawn from the existing literature (potentially including authors' own previously published work) are accompanied by appropriate citations}{(yes/no/NA)}
	yes

	\question{All datasets drawn from the existing literature (potentially including authors' own previously published work) are publicly available}{(yes/partial/no/NA)}
	yes

	\question{All datasets that are not publicly available are described in detail, with explanation why publicly available alternatives are not scientifically satisficing}{(yes/partial/no/NA)}
	yes

\end{itemize}
\end{itemize}

\checksubsection{Computational Experiments}
\begin{itemize}

\question{Does this paper include computational experiments?}{(yes/no)}
yes

\ifyespoints{If yes, please address the following points:}
\begin{itemize}

	\question{This paper states the number and range of values tried per (hyper-) parameter during development of the paper, along with the criterion used for selecting the final parameter setting}{(yes/partial/no/NA)}
	yes

	\question{Any code required for pre-processing data is included in the appendix}{(yes/partial/no)}
	partial

	\question{All source code required for conducting and analyzing the experiments is included in a code appendix}{(yes/partial/no)}
	yes

	\question{All source code required for conducting and analyzing the experiments will be made publicly available upon publication of the paper with a license that allows free usage for research purposes}{(yes/partial/no)}
	partial
        
	\question{All source code implementing new methods have comments detailing the implementation, with references to the paper where each step comes from}{(yes/partial/no)}
	yes

	\question{If an algorithm depends on randomness, then the method used for setting seeds is described in a way sufficient to allow replication of results}{(yes/partial/no/NA)}
	yes

	\question{This paper specifies the computing infrastructure used for running experiments (hardware and software), including GPU/CPU models; amount of memory; operating system; names and versions of relevant software libraries and frameworks}{(yes/partial/no)}
	partial

	\question{This paper formally describes evaluation metrics used and explains the motivation for choosing these metrics}{(yes/partial/no)}
	partial

	\question{This paper states the number of algorithm runs used to compute each reported result}{(yes/no)}
	yes

	\question{Analysis of experiments goes beyond single-dimensional summaries of performance (e.g., average; median) to include measures of variation, confidence, or other distributional information}{(yes/no)}
	no

	\question{The significance of any improvement or decrease in performance is judged using appropriate statistical tests (e.g., Wilcoxon signed-rank)}{(yes/partial/no)}
	partial

	\question{This paper lists all final (hyper-)parameters used for each model/algorithm in the paper’s experiments}{(yes/partial/no/NA)}
	yes

\end{itemize}
\end{itemize}
\ifreproStandalone
\end{document}
\fi


\end{document}